\documentclass[runningheads]{llncs}
\usepackage[T1]{fontenc}
\usepackage{graphicx,verbatim}
\usepackage{hyperref}
\usepackage{color}

\usepackage{array}
\usepackage{tabularx}
\usepackage{pgfplots}
\usepgfplotslibrary{groupplots}
\pgfplotsset{compat=1.18}
\usepackage{multirow}
\usepackage{xcolor} 
\newcolumntype{Y}{>{\centering\arraybackslash}X}

\usepackage{tikz}
\usetikzlibrary{shapes.geometric, arrows.meta, positioning, shadows, calc}

\definecolor{sciBlue}{RGB}{225, 235, 245}
\definecolor{sciRed}{RGB}{250, 230, 230}
\definecolor{sciGreen}{RGB}{230, 245, 230}
\definecolor{sciYellow}{RGB}{255, 252, 225}
\definecolor{sciPurple}{RGB}{240, 230, 250}
\definecolor{sciText}{RGB}{30, 30, 30}
\definecolor{sciDraw}{RGB}{70, 70, 70}

\begin{document}
\title{CytoSyn: a Foundation Diffusion Model for Histopathology - Tech Report}
%
\author{Thomas Duboudin\textsuperscript{$\dagger$} \and
Xavier Fontaine \and Etienne Andrier \and Lionel Guillou \and Alexandre Filiot \and Thalyssa Baiocco-Rodrigues \and Antoine Olivier \and Alberto Romagnoni \and John Klein \and Jean-Baptiste Schiratti}
\authorrunning{T. Duboudin, X. Fontaine, E. Andrier et al.}
%
\institute{Owkin, Inc \\
\textsuperscript{$\dagger$} Corresponding author \\
\email{firstname.lastname@owkin.com}}

\maketitle              
\begin{abstract}

Computational pathology has made significant progress in recent years, fueling advances in both fundamental disease understanding and clinically ready tools. This evolution is driven by the availability of large amounts of digitized slides and specialized deep learning methods and models. Multiple self-supervised foundation feature extractors have been developed, enabling downstream predictive applications from cell segmentation to tumor sub-typing and survival analysis. In contrast, generative foundation models designed specifically for histopathology remain scarce. Such models could address tasks that are beyond the capabilities of feature extractors, such as virtual staining. In this paper, we introduce CytoSyn, a state-of-the-art foundation latent diffusion model that enables the guided generation of highly realistic and diverse histopathology H\&E-stained images, as shown in an extensive benchmark. We explored methodological improvements, training set scaling, sampling strategies and slide-level overfitting, culminating in the improved CytoSyn-v2, and compared our work to PixCell, a state-of-the-art model, in an in-depth manner. This comparison highlighted the strong sensitivity of both diffusion models and performance metrics to preprocessing-specific details such as JPEG compression. Our model has been trained on a dataset obtained from more than 10,000 TCGA diagnostic whole-slide images of 32 different cancer types. Despite being trained only on oncology slides, it maintains state-of-the-art performance generating inflammatory bowel disease images. To support the research community, we publicly release CytoSyn's weights, its training and validation datasets, and a sample of synthetic images in this repository: \href{https://huggingface.co/Owkin-Bioptimus/CytoSyn}{https://huggingface.co/Owkin-Bioptimus/CytoSyn}.

\keywords{Image synthesis \and Digital pathology \and Diffusion \and CytoSyn}

\end{abstract}

\section{Introduction}

Most modern computational pathology pipelines are built upon  large deep-learning models trained in a self-supervised (SSL) fashion to extract semantically-rich features from pathology images. SSL foundation backbones have outperformed models trained on labeled datasets by leveraging a significantly higher amount of data and larger model size. Several such models~\cite{hoptimus0,chen2024towards,zimmermann2024virchow2,lu2024visual,filiot2024phikon} have been created specifically for digital pathology (a field in which annotated data is scarce) and enable a wide range of downstream predictive applications: tissue and cells segmentation~\cite{histoplus2025}, gene expression prediction~\cite{jaume2024hest}, tumor sub-typing and survival analysis~\cite{neidlinger2025benchmarking,gatopoulos2024eva,campanella2025clinical}, etc. These applications allow researchers to both build clinically usable tools and derive deep biological insights. However, SSL foundation models are not designed to effectively address all questions of interest to the computational pathology field. We argue that a domain-specific image generation model would be helpful to better tackle some of these problems. For instance, they are not easily interpretable with interpretation usually only happening with regard to a particular downstream task using attention scores or Shapley values. Generative models offer a path toward counterfactual interpretability, allowing researchers to visualize how an image would change if specific features were missing or over-amplified. Feature extractors also cannot perform inherently generative tasks, such as virtual staining: an approach used to mitigate performance degradation due to scanner and staining variability that relies on being able to transfer staining while keeping the biologically-relevant content unchanged. Furthermore, standard data augmentation cannot counteract the lack of diversity in rare diseases or tissue types datasets. This could be addressed by a generative model going beyond simple geometric transformations.\\

\begin{figure}[ht]
    \centering
    \includegraphics[width=\linewidth]{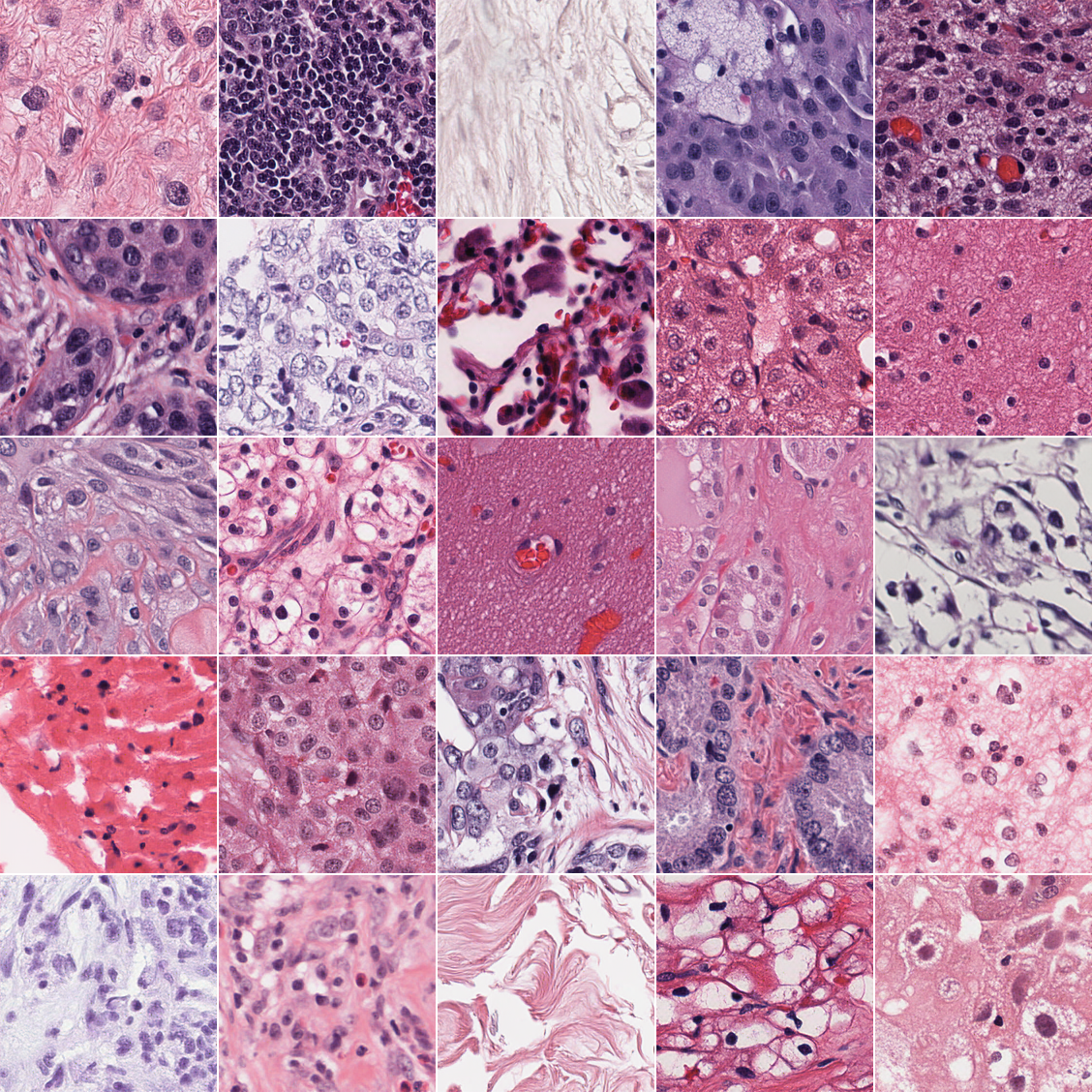}
    \caption{Examples of tiles generated unconditionally with CytoSyn.}
    \label{fig:unconditional}
\end{figure}

Diffusion and flow matching models have become the de facto standard for image synthesis in recent years. However, most of the publicly available models are trained for illustrative, graphic design, or photo editing purposes on "natural" image datasets. They are thus unsuitable to generate highly-specific images such as H\&E histopathology images, in which fine details (such as the shapes, types and organization of cells) can contain a lot of biologically relevant information. In this paper, we therefore introduce CytoSyn, a foundational diffusion model specifically tailored to generate H\&E-stained pathology images that is able to generate highly realistic and diversified images (a sample of generated images is available in Figure \ref{fig:unconditional}).\\

Our contributions in this paper are threefold: 
\begin{itemize}
    \item We built CytoSyn, a state-of-the-art diffusion model. Building upon REPA-E~\cite{repae}, we introduce some methodological novelties to tailor the architecture to histopathology.
    \item We benchmarked it extensively, including an out-of-distribution scenario on non-oncology tissue, and performed an in-depth comparison against the current state-of-the-art model, revealing the high impact of the data preparation steps on the final output.
    \item We publicly release both the model weights and the data used to train and benchmark it to support the research community and to ensure reproducibility.
\end{itemize}

\newpage

\section{Related Works}

\subsection{Diffusion models}

In the last few years, Generative Adversarial Networks (GANs) have been outperformed by diffusion models~\cite{sohl2015deep,ho2020denoising,dhariwal2021diffusion} and score-based generative models~\cite{song2019generative} to produce high-quality images. Diffusion models work by gradually adding noise to input data and learning the backward denoising process. Score-based methods work by producing samples using a Langevin dynamics after estimation of the score $\nabla \log p(x)$. These two approaches have been unified by Song et al.~\cite{song2020score} who showed that the reverse diffusion process can be modeled by a stochastic differential equation containing the score of the data distribution. Many improvements have been designed over the initial diffusion models including guidance~\cite{dhariwal2021diffusion,ho2021classifier}, replacing the original U-Net architecture by a vision transformer~\cite{dit} and the use of the latent space of a VAE to perform the diffusion process~\cite{ldm}, allowing the faster generation of larger images with a limited computational power. Further methods have been then proposed to improve training efficiency and quality generation, among them REPA~\cite{repa} and REPA-E~\cite{repae} which use an additional feature extractor to align the hidden state representations of the denoising network with the embeddings of the input data.\\

Flow matching is another state-of-the-art technique for generative modeling~\cite{lipman2022flow,liu2022flow} which is actually equivalent to diffusion models~\cite{gao2025diffusionmeetsflow}. The Stochastic Interpolants Framework~\cite{albergo2023stochastic} unify both approaches with a general formulation that allows more flexible paths from the noise to the data distribution, as well as different sampling options. The SiT model~\cite{sit} builds upon this work and proposes improvements over a classical Diffusion Transformer by using notably stochastic sampling (from the SDE) instead of deterministic sampling (from the ODE), which improves the quality of the generated images despite requiring a higher computational budget.

\subsection{Diffusion applied to digital pathology}

Diffusion-based image synthesis has emerged recently within the computational pathology community but has already been explored for a wide range of purposes: virtual staining~\cite{tsai2024test}, improving self-supervised foundation models and downstream predictive models~\cite{graikos2024learned,belagali2024gen}, enabling privacy-preserving~\cite{yellapragada2025pixcell} or interpretability~\cite{vzigutyte2025counterfactual} applications, and generating whole-slide images~\cite{Yellapragada_2025_CVPR} (as opposed to tile-level synthesis). Another line of research bridges histology with transcriptomic data by conditioning generation on RNA expression profiles~\cite{carrillo2025generation}. However, until recently, most approaches trained their own backbone diffusion models on limited amounts of data, on select indications, or with a highly specific conditioning mechanism, thereby preventing them from generalizing beyond their originally envisioned applications (e.g. tumor or non-tumor binary labels that are meaningless in a non-oncology setting). 

\subsection{PixCell}

To the best of our knowledge, PixCell~\cite{yellapragada2025pixcell} is currently the only other publicly released foundation diffusion model for histopathology. Our work most closely resembles the base model PixCell-256 but several architectural and methodological distinctions exist. Primarily, our approach is based on REPA-E and enforce representation alignment during training, whereas PixCell follows the conventional Latent Diffusion Model (LDM) approach with a frozen VAE and no specific training constraints in addition to the standard reconstruction loss.\\

From an architecture perspective, the models differ notably in their choice of conditioning model: CytoSyn employs H0-mini (86M parameters) for guidance while PixCell uses UNI2-h~\cite{chen2024towards} (680M parameters), making CytoSyn's VRAM requirements at inference time lower. Furthermore, PixCell utilizes a frozen VAE from Stable Diffusion v3~\cite{esser2024scaling} (SD3.5 Large) trained on natural images while we trained our VAE from scratch on histopathology data, with the goal of learning better pathology-specific features. Finally, we only trained our model on TCGA diagnostic slides as we envisioned oncology-focused predictive applications. We therefore excluded the GTEX slides from healthy samples and the fresh frozen TCGA slides as they are usually not suitable for these purposes. In contrast, PixCell's training set is more diverse with data coming from a mix of TCGA (both diagnostic and fresh frozen), GTEX~\cite{lonsdale2013genotype}, CPTAC~\cite{edwards2015cptac} and other sources. In Table \ref{table:pixcell_vs_cytosyn} we summarize all differences between the two models. We compare them and explore the impact of some of these choices in the Experiments section. In the rest of the paper, by PixCell we denote the PixCell-256 model, not its PixCell-1024 counterpart.

\begin{table}[]
    \centering
    \caption{Differences between PixCell and CytoSyn}
    \label{table:pixcell_vs_cytosyn}
    \begin{tabular}{c|c|c}
        Model & PixCell-256 & CytoSyn\\
        \hline
        Framework & Standard LDM & REPA-E \\
        Diffusion Model & DiT-XL/2 & SiT-XL/2 \\
        Sampling scheme & DPM-Solver & Euler-Maruyama \\
        VAE & SD3.5 Large VAE, frozen & SD-VAE, f8d4, trained\\
        Conditioning & UNI2-h (ViT-h/14) & H0-mini (ViT-B/14) \\
        Data Sources & GTEX, CPTAC, TCGA \& others & TCGA diag. \\
        \# Tiles in training set  & $\sim 31M$ & $\sim 40M$ / $\sim 108M$ \\
        \# Slides in training set & $\sim 69k$ & $\sim 10.6k$ \\
        Image size & $256 \times 256$ & $224 \times 224$ \\
        Tiling pipeline & DS-MIL & Internal \\
    \end{tabular}

\end{table}

\section{Method}

\subsection{Architecture}

\begin{figure}[h!]
    \centering
    \includegraphics[width=\linewidth]{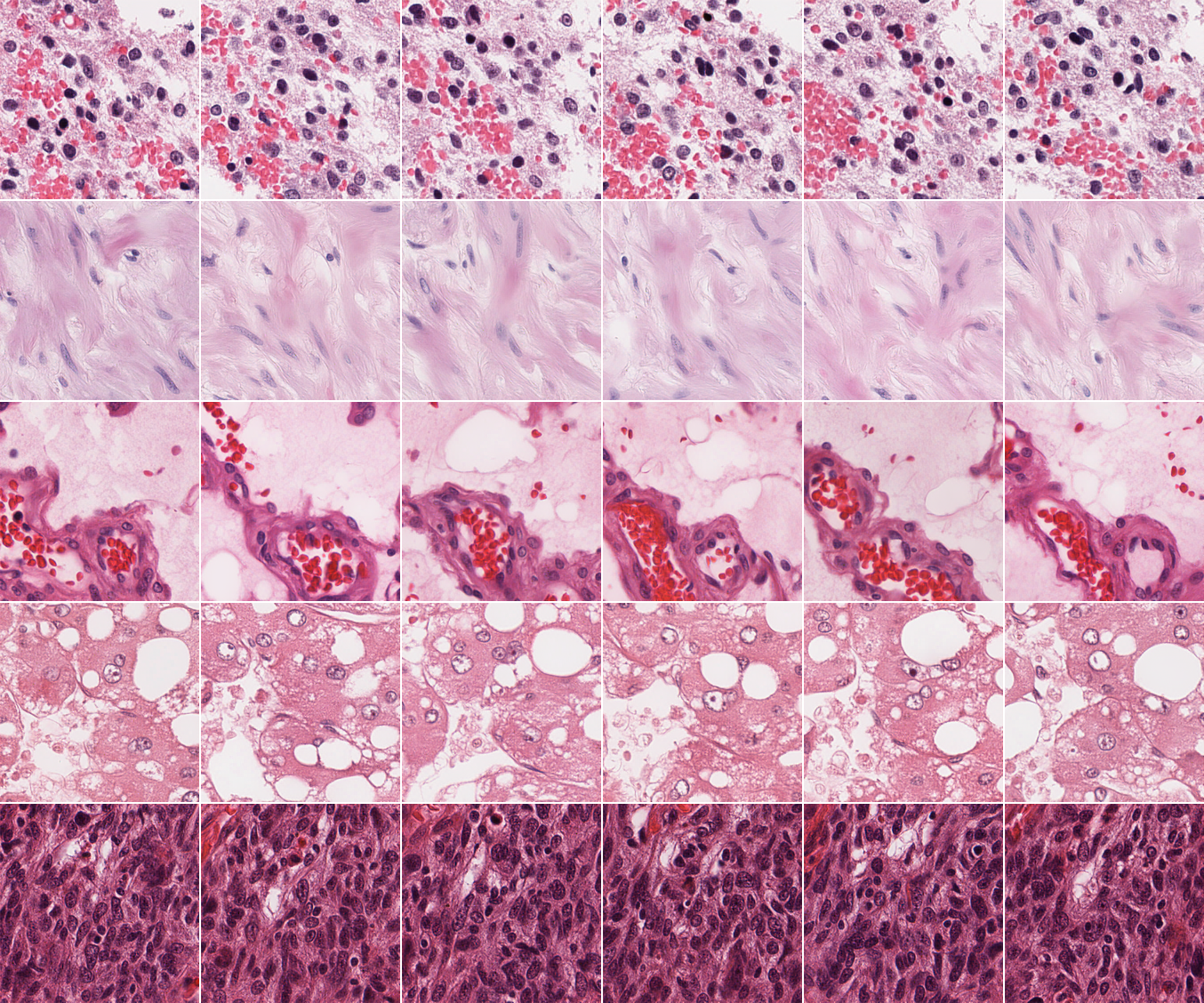}
    \caption{H0-mini conditioning enables the generation of visually distinct yet biologically highly consistent tiles. Each row shows one reference image (left) and five generated variations.}
    \label{fig:conditional}
\end{figure}

\begin{figure}[h!]
    \centering
    \includegraphics[width=\linewidth]{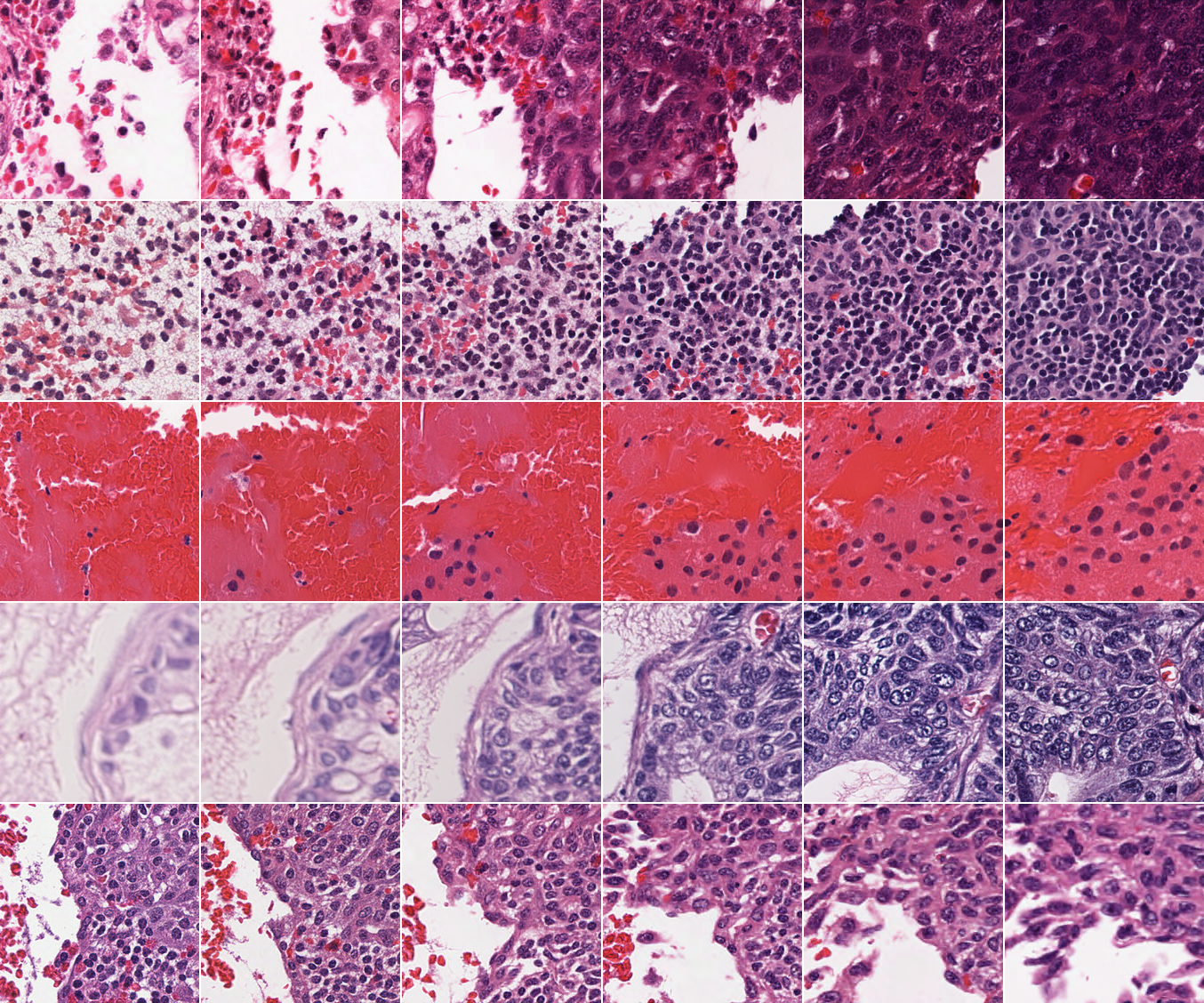}
    \caption{Feature-based conditioning allows a fine-grained control on the semantic of the synthesized images, a prerequisite to use synthetic images as data augmentation, while maintaining highly realistic outputs as illustrated in this figure with a linear interpolation example. Left and right columns: original tiles, center columns: synthetic tiles obtained using a linear interpolation of left and right tiles' features (with interpolation factor $0.2, 0.4, 0.6, 0.8$).}
    \label{fig:interpolation}
\end{figure}

CytoSyn is based on the REPA-E architecture~\cite{repae}, itself a modification of REPA~\cite{repa}. The REPA architecture is a latent-diffusion architecture~\cite{ldm} (LDM) with an additional alignment constraint: the patch tokens of the diffusion transformer model are aligned to those of a frozen self-supervised transformer using a cosine similarity loss. It was found to make training much faster and improve the quality of generated images. REPA-E builds upon REPA by training both the VAE and the diffusion model at the same time: in REPA, the VAE is trained beforehand and frozen during the training of the diffusion model, whereas in REPA-E, the two models are trained simultaneously with specific care taken to avoid collapse. This yielded additional gains in both training speed and generation quality. We introduced several modifications compared to the original REPA-E:
\begin{itemize}
    \item \textbf{Image size}: The default size for generated images is $256 \times 256$ for both REPA and REPA-E. In the computational pathology field, for legacy reasons, most feature extractors expect as inputs images of $224 \times 224$ pixels. We therefore decided to generate images at this particular dimension to ease further processing: no need for additional image resizing or cropping.
    \item \textbf{Representation alignment}: The original REPA and REPA-E methods use the ViT-B/14 extractor from DINOv2~\cite{oquab2024dinov2}. DINOv2 models are near state-of-the-art SSL feature extractors trained on a curated subset of the LVD-142M dataset, which contains ImageNet-like images that are very different from histopathology images. We therefore replaced the DINOv2 model with a publicly available feature extractor trained on histopathology data: H0-mini~\cite{filiot2025distilling}. We chose this SSL model among many as it achieves high performance on many downstream tasks, indicating a good capacity at extracting informative and generalist embeddings, while still being lightweight (ViT-B/14). For an image of size $224 \times 224$, H0-mini yields $16 \times 16$ patch tokens, and the f8d4 SD-VAE~\cite{ldm} a latent of size $28 \times 28$. When this latent is sent through the SiT-XL/2~\cite{sit} diffusion model, we get $14 \times 14$ patch tokens. To enable the alignment of the tokens, we opted to subsample the spatialized H0-mini tokens to $14 \times 14$ with a bicubic interpolation and anti-aliasing instead of doing the opposite (upsampling the $14 \times 14$ SiT-XL/2 tokens), allowing for the pre-computing of H0-mini resized patch tokens.
    \item \textbf{Conditioning}: REPA-E and REPA rely on classifier-free guidance~\cite{ho2021classifier} to enable the synthesis of images based on additional semantic information (such as a caption, a label, or a semantic segmentation map). We opted to use SSL features to encompass the semantic information present in a tile, as in PixCell, due to the lack of large-scale datasets with fine-grained tile-level annotations. We hypothesized that slide-level labels (such as the indication) lacked the granularity required to be a useful supervisory signal. We again used H0-mini for the tile-level conditioning ([CLS]-token for the conditioning, patch tokens for the alignment). We did not explore different pairs of SSL models for guidance and alignment, as using a single model is more computationally efficient: a single forward pass yields both the alignment tokens and the guidance token. Samples generated conditionally with H0-mini guidance are available in Figure \ref{fig:conditional}. Feature-based guidance enables a fine-grained control on the semantic of the generated images that text-based guidance does not (as can be seen in Figure \ref{fig:interpolation}).
    \item \textbf{REPA post-training}: REPA-E shows that the best generation results can be achieved by first training end-to-end the full architecture, and then using the obtained VAE (in a frozen fashion) to train a diffusion model with the REPA architecture. Due to computational limitations, we did not perform this additional step and all the results of the paper are from end-to-end trainings.
    \item \textbf{Initialization}: In REPA-E, the VAE weights at initialization are those of an already trained VAE (e.g SD-VAE, VA-VAE). In our case, given the specificity of histopathology data and its overall abundance (in a non-annotated format) we trained all the models from scratch.
    \item \textbf{VAE EMA}: It has been shown that computing an exponential moving average (EMA) of both the latent diffusion model and the VAE was beneficial to performance and this has since become common practice~\cite{ldm,ho2020denoising,dhariwal2021diffusion}. We computed such an EMA of the VAE during training to be used at inference-time, as an EMA model is computed only for the latent diffusion and not for the VAE in the original REPA-E paper (we used the exact same EMA parameters for both models). Whether the raw VAE model or the EMA version is used will be indicated in the results.
\end{itemize}

The model therefore consists of 3 different components, totaling 853M parameters, out of which 767M are trained:
\begin{itemize}
    \item a variational auto-encoder (SD-VAE, f8d4 version, 84M parameters),
    \item a transformer latent diffusion model (SiT XL/2, 683M parameters),
    \item H0-mini, used frozen as both the guidance and the representation alignment model (ViT-b/14, 86M parameters).
\end{itemize}

\subsection{Dataset}


Histopathology slides (also called whole-slide images) are usually digitized via very high resolution scanning, resulting in very large images that cannot be processed entirely at once by any deep computer vision model. Slides are therefore partitioned into collections of smaller images, called tiles, extracted from the areas containing tissue and liekly devoid of artifacts (e.g., folds, bubbles, pen marks, out-of-focus areas, and dust). To perform this operation, we use a proprietary pipeline, built upon a tissue detection model, that ingests the slides at lower resolution and excludes empty spaces and artifacts from the extraction. Using this pipeline, we extracted 40M ($224 \times 224$) randomly sampled tiles from 10,622 TCGA \cite{weinstein2013cancer} diagnostic slides at 0.5 microns per pixel (MPP, equivalent to $20\times$ magnification). TCGA slides have a tissue source site (TSS), which is a code for both the hospital or the research center from which the tissue samples have been sourced and the indication. Given that all centers use potentially different scanners and staining protocols, we sampled our 40M tiles to ensure a stratified representation of TSS codes, mirroring the global TCGA distribution. This encompassed images from 32 different indications (and 679 TSS).\\

Additionally, to investigate scaling behavior, we created an expanded training set. Starting with 11,520 TCGA diagnostic H\&E whole-slide images across 32 indications, we applied a curation process to remove artifacted tiles, yielding a curated dataset comprising 115M tiles. Artifact curation was performed using an in-house ViT-Small model pre-trained with iBOT~\cite{zhou2022image} on TCGA-COAD (3.9M tiles), incorporating histology-specific augmentations~\cite{shen2022randstainna}. Using the frozen backbone features, a linear classifier was trained under 5-fold cross-validation on 79.5k tile-level annotations distinguishing usable tissue from artifacts. At inference, predictions were averaged across folds. This procedure removed 1.6M tiles (1.4\%). Finally, we removed all necessary validation tiles to prevent data leakage, resulting in a total of 108M tiles.

\subsection{Training}

The models have been trained on 64 A100 GPUs with a total batch size of 640. Other training parameters were kept to their default value in the REPA-E repository (unless specified otherwise). In particular, the classifier-free guidance scale is set to 2.5, and the guidance-high (resp. -low) parameter is set to 0.75 (resp. 0). The entirety of the experiments detailed in the paper represents around 40k GPU-hours.

\section{Experiments}

\subsection{Benchmark}

To rigorously benchmark our model, we created 2 validation sets of 100k images using the TCGA cohort, stratified to maintain the TSS distribution. These sets differ based on their source slides: in both cases, the selected tiles are non-overlapping with the training tiles. However, for the \textbf{val-in} dataset, the tiles originate from slides from which some tiles were extracted for use in training. In the \textbf{val-out} dataset, the tiles' originating slides are entirely distinct from the training set's slides. We held out 1,012 slides for the \textbf{val-out} dataset, covering 32 indications and 359 TSS. These two datasets will enable us to assess the level of slide-level overfitting in histopathology-specific generative models by comparing results. We generated 100k images with each benchmarked model, using features from 100k tiles randomly sampled from the training set (stratified by TSS) as guidance, following PixCell's methodology. Main results have been obtained with 250 steps of SDE sampling and all models have been trained over the same number of epochs.\\

\begin{table}[ht]
\centering
\fontsize{8}{\baselineskip}\selectfont
\caption{Performance comparison of CytoSyn models across different metrics and feature extractors (\textbf{val-in} / \textbf{val-out} values in the table cells).}
\label{tab:model_comparison}
\begin{tabular}{|l|l|c|c|c|c|}
\hline
Metric & Model & H-Optimus-0 & Virchow 2 & UNI2-h & Inception V3 \\ \hline
\multirow{3}{*}{FD} & CytoSyn - 40M & 58.4 / 72.2 & 55.3 / 70.6 & 10.9 / 16.7 & \textbf{2.9} / \textbf{3.4} \\
 & CytoSyn - 108M & 58.4 / 72.3 & 56.8 / 71.5 & 12.5 / 18.3 & 3.7 / 4.1 \\
 & CytoSyn - 108M - EMA & \textbf{48.1} / \textbf{62.5} & \textbf{50.1} / \textbf{63.5} & \textbf{9.4} / \textbf{15.1} & 3.4 / 3.9 \\ \hline
 \hline
 FD & Guidance vs Val. sets & 4.0 / 20.1 & 3.5 / 20.6 & 1.4 / 7.7 & 0.5 / 0.8 \\
 \hline\hline
\multirow{3}{*}{FLD} & CytoSyn - 40M & 11.4 / 10.4 & \textbf{3.4} / \textbf{3.9} & 9.0 / 4.9 & 1.9 / \textbf{0.5} \\
 & CytoSyn - 108M & \textbf{11.1} / \textbf{9.7} & 4.1 / 3.6 & 6.3 / \textbf{4.8} & \textbf{1.0} / 4.5 \\
 & CytoSyn - 108M - EMA & 11.6 / 10.6 & 4.3 / 4.0 & \textbf{6.2} / 4.9 & 2.6 / 3.2 \\ \hline
\multirow{3}{*}{Precision} & CytoSyn - 40M & 0.94 / 0.94 & 0.95 / 0.95 & 0.98 / 0.98 & 0.82 / 0.82 \\
 & CytoSyn - 108M & 0.95 / 0.95 & 0.96 / 0.96 & 0.98 / 0.98 & 0.83 / 0.83 \\
 & CytoSyn - 108M - EMA & \textbf{0.96} / \textbf{0.96} & \textbf{0.96} / \textbf{0.96} & \textbf{0.98} / \textbf{0.98} & \textbf{0.83} /\textbf{ 0.83} \\ \hline
\multirow{3}{*}{Recall} & CytoSyn - 40M & 0.99 / 0.99 & 0.99 / 0.99 & 0.98 / 0.98 & 0.89 / 0.89 \\
 & CytoSyn - 108M & 0.99 / 0.99 & 0.99 / 0.99 & 0.97 / 0.97 & 0.88 / 0.88 \\
 & CytoSyn - 108M - EMA & \textbf{0.99} / \textbf{0.99} & \textbf{0.99} / \textbf{0.99} & \textbf{0.99} / \textbf{0.99} & \textbf{0.90} / \textbf{0.90} \\ \hline
\multirow{3}{*}{Cosine Sim} & CytoSyn - 40M & 0.78 / 0.78 & 0.90 / 0.90 & 0.79 / 0.78 & 0.88 / 0.88 \\
 & CytoSyn - 108M & 0.79 / 0.79 & 0.91 / 0.91 & 0.79 / 0.79 & 0.88 / 0.88 \\
 & CytoSyn - 108M - EMA & \textbf{0.80} / \textbf{0.80} & \textbf{0.91} / \textbf{0.91} & \textbf{0.80} / \textbf{0.80} & \textbf{0.88} / \textbf{0.88} \\ \hline

\end{tabular}
\end{table}

\begin{figure*}[h!]
    \centering
    \resizebox{\textwidth}{!}{%
    \begin{tikzpicture}

    \begin{scope}[shift={(0cm,5.5cm)}]
    \begin{axis}[
        width=7.8cm, height=6cm,
        legend pos=north east, ymajorgrids=true, grid style=dashed,
        xmin=0, xmax=270, tick label style={font=\footnotesize},
        legend style={font=\fontsize{8}{9.6}\selectfont},
        title={Inception V3},
        name=main_plot0
    ]
        \addplot[color=red, mark=*, thick] coordinates {(20,39.5) (50,11.0) (250,5.2)};
        \addlegendentry{ODE (val-out)}
        \addplot[color=blue, mark=square*, thick] coordinates {(20,30.4) (50,7.9) (250,5.4)};
        \addlegendentry{SDE (val-out)}
        \addplot[color={rgb,255:red,34;green,139;blue,34}, mark=*, thick, dashed] coordinates {(20,39.6) (50,10.7) (250,4.8)};
        \addlegendentry{ODE (val-in)}
        \addplot[color={rgb,255:red,218;green,165;blue,32}, mark=square*, thick, dashed] coordinates {(20,30.2) (50,7.5) (250,4.9)};
        \addlegendentry{SDE (val-in)}
        \draw [draw=black, dashed, thin] (axis cs:240,4.4) rectangle (axis cs:260,6.0);
    \end{axis}

    \begin{axis}[
        at={(main_plot0.south east)}, xshift=-0.4cm, yshift=0.8cm, anchor=south east,
        width=3.0cm, height=3.0cm,
        xmin=240, xmax=265, ymin=4.4, ymax=6.0,
        tick label style={font=\tiny}, xtick={240, 260}, ytick={4.4, 6.0},
        grid=both, grid style={line width=.1pt, draw=gray!20},
    ]
        \addplot[color=red, mark=*, thick] coordinates {(250,5.2)} node[right, xshift=2pt, font=\tiny, color=red] {5.2};
        \addplot[color=blue, mark=square*, thick] coordinates {(250,5.4)} node[right, xshift=2pt, font=\tiny, color=blue] {5.4};
        \addplot[color={rgb,255:red,34;green,139;blue,34}, mark=*, thick, dashed] coordinates {(250,4.8)} node[right, xshift=2pt, font=\tiny, color={rgb,255:red,34;green,139;blue,34}] {4.8};
        \addplot[color={rgb,255:red,218;green,165;blue,32}, mark=square*, thick, dashed] coordinates {(250,4.9)} node[right, xshift=2pt, font=\tiny, color={rgb,255:red,218;green,165;blue,32}] {4.9};
    \end{axis}
    \end{scope}

    \begin{scope}[shift={(7.3cm,5.5cm)}]
    \begin{axis}[
        width=7.8cm, height=6cm,
        ymajorgrids=true, grid style=dashed,
        xmin=0, xmax=270, tick label style={font=\footnotesize},
        title={H-Optimus-0},
        name=main_plot1
    ]
        \addplot[color=red, mark=*, thick] coordinates {(20,282.2) (50,192.5) (250,171.3)};
        \addplot[color=blue, mark=square*, thick] coordinates {(20,239.0) (50,183.3) (250,170.7)};
        \addplot[color={rgb,255:red,34;green,139;blue,34}, mark=*, thick, dashed] coordinates {(20,273.5) (50,181.2) (250,159.6)};
        \addplot[color={rgb,255:red,218;green,165;blue,32}, mark=square*, thick, dashed] coordinates {(20,226.5) (50,171.3) (250,158.7)};
        \draw [draw=black, dashed, thin] (axis cs:240,154.0) rectangle (axis cs:260,176.0);
    \end{axis}

    \begin{axis}[
        at={(main_plot1.south east)}, xshift=-0.4cm, yshift=1.6cm, anchor=south east,
        width=3.0cm, height=3.0cm,
        xmin=240, xmax=265, ymin=154.0, ymax=176.0,
        tick label style={font=\tiny}, xtick={240, 260}, ytick={154.0, 176.0},
        grid=both, grid style={line width=.1pt, draw=gray!20},
    ]
        \addplot[color=red, mark=*, thick] coordinates {(250,171.3)} 
            node[right, xshift=2pt, yshift=3pt, font=\tiny, color=red] {171.3};
        \addplot[color=blue, mark=square*, thick] coordinates {(250,170.7)} 
            node[right, xshift=2pt, yshift=-3pt, font=\tiny, color=blue] {170.7};
        \addplot[color={rgb,255:red,34;green,139;blue,34}, mark=*, thick, dashed] coordinates {(250,159.6)} 
            node[right, xshift=2pt, yshift=3pt, font=\tiny, color={rgb,255:red,34;green,139;blue,34}] {159.6};
        \addplot[color={rgb,255:red,218;green,165;blue,32}, mark=square*, thick, dashed] coordinates {(250,158.7)} 
            node[right, xshift=2pt, yshift=-3pt, font=\tiny, color={rgb,255:red,218;green,165;blue,32}] {158.7};
    \end{axis}
    \end{scope}

    \begin{scope}[shift={(0cm,0cm)}]
    \begin{axis}[
        width=7.8cm, height=6cm,
        ymajorgrids=true, grid style=dashed,
        xmin=0, xmax=270, tick label style={font=\footnotesize},
        title={UNI2-h},
        name=main_plot2
    ]
        \addplot[color=red, mark=*, thick] coordinates {(20,68.3) (50,45.5) (250,40.2)};
        \addplot[color=blue, mark=square*, thick] coordinates {(20,57.7) (50,42.9) (250,39.1)};
        \addplot[color={rgb,255:red,34;green,139;blue,34}, mark=*, thick, dashed] coordinates {(20,64.4) (50,41.1) (250,35.5)};
        \addplot[color={rgb,255:red,218;green,165;blue,32}, mark=square*, thick, dashed] coordinates {(20,52.7) (50,38.2) (250,34.5)};
        \draw [draw=black, dashed, thin] (axis cs:240,32.0) rectangle (axis cs:260,43.0);
    \end{axis}

    \begin{axis}[
        at={(main_plot2.south east)}, xshift=-0.4cm, yshift=1.6cm, anchor=south east,
        width=3.0cm, height=3.0cm,
        xmin=240, xmax=265, ymin=32.0, ymax=43.0,
        tick label style={font=\tiny}, xtick={240, 260}, ytick={32.0, 43.0},
        grid=both, grid style={line width=.1pt, draw=gray!20},
    ]
        \addplot[color=red, mark=*, thick] coordinates {(250,40.2)} 
            node[right, xshift=2pt, font=\tiny, color=red] {40.2};
        \addplot[color=blue, mark=square*, thick] coordinates {(250,39.1)} 
            node[right, xshift=2pt, font=\tiny, color=blue] {39.1};
        \addplot[color={rgb,255:red,34;green,139;blue,34}, mark=*, thick, dashed] coordinates {(250,35.5)} 
            node[right, xshift=2pt, font=\tiny, color={rgb,255:red,34;green,139;blue,34}] {35.5};
        \addplot[color={rgb,255:red,218;green,165;blue,32}, mark=square*, thick, dashed] coordinates {(250,34.5)} 
            node[right, xshift=2pt, font=\tiny, color={rgb,255:red,218;green,165;blue,32}] {34.5};
    \end{axis}
    \end{scope}

    \begin{scope}[shift={(7.3cm,0cm)}]
    \begin{axis}[
        width=7.8cm, height=6cm,
        ymajorgrids=true, grid style=dashed,
        xmin=0, xmax=270, tick label style={font=\footnotesize},
        title={Virchow 2},
        name=main_plot3
    ]
        \addplot[color=red, mark=*, thick] coordinates {(20,372.7) (50,228.8) (250,203.7)};
        \addplot[color=blue, mark=square*, thick] coordinates {(20,311.4) (50,214.0) (250,197.4)};
        \addplot[color={rgb,255:red,34;green,139;blue,34}, mark=*, thick, dashed] coordinates {(20,363.5) (50,217.6) (250,191.5)};
        \addplot[color={rgb,255:red,218;green,165;blue,32}, mark=square*, thick, dashed] coordinates {(20,297.5) (50,201.6) (250,185.8)};
        \draw [draw=black, dashed, thin] (axis cs:240,178.0) rectangle (axis cs:260,212.0);
    \end{axis}

    \begin{axis}[
        at={(main_plot3.south east)}, xshift=-0.4cm, yshift=1.6cm, anchor=south east,
        width=3.0cm, height=3.0cm,
        xmin=240, xmax=265, ymin=178.0, ymax=212.0,
        tick label style={font=\tiny}, xtick={240, 260}, ytick={178.0, 212.0},
        grid=both, grid style={line width=.1pt, draw=gray!20},
    ]
        \addplot[color=red, mark=*, thick] coordinates {(250,203.7)} 
            node[right, xshift=2pt, font=\tiny, color=red] {203.7};
        \addplot[color=blue, mark=square*, thick] coordinates {(250,197.4)} 
            node[right, xshift=2pt, font=\tiny, color=blue] {197.4};
        \addplot[color={rgb,255:red,34;green,139;blue,34}, mark=*, thick, dashed] coordinates {(250,191.5)} 
            node[right, xshift=2pt, font=\tiny, color={rgb,255:red,34;green,139;blue,34}] {191.5};
        \addplot[color={rgb,255:red,218;green,165;blue,32}, mark=square*, thick, dashed] coordinates {(250,185.8)} 
            node[right, xshift=2pt, font=\tiny, color={rgb,255:red,218;green,165;blue,32}] {185.8};
    \end{axis}
    \end{scope}

    \end{tikzpicture}
    }
    \caption{Unconditional image generation performance of CytoSyn (40M model) across different feature extractors, number of sampling steps, sampling methods and validation sets ($y$-axis: Fréchet distance, $x$-axis: number of sampling steps). The inset box in each plot provides a magnified view of the values obtained with 250 sampling steps.}
    \label{fig:uncond_validation_grid}
\end{figure*}
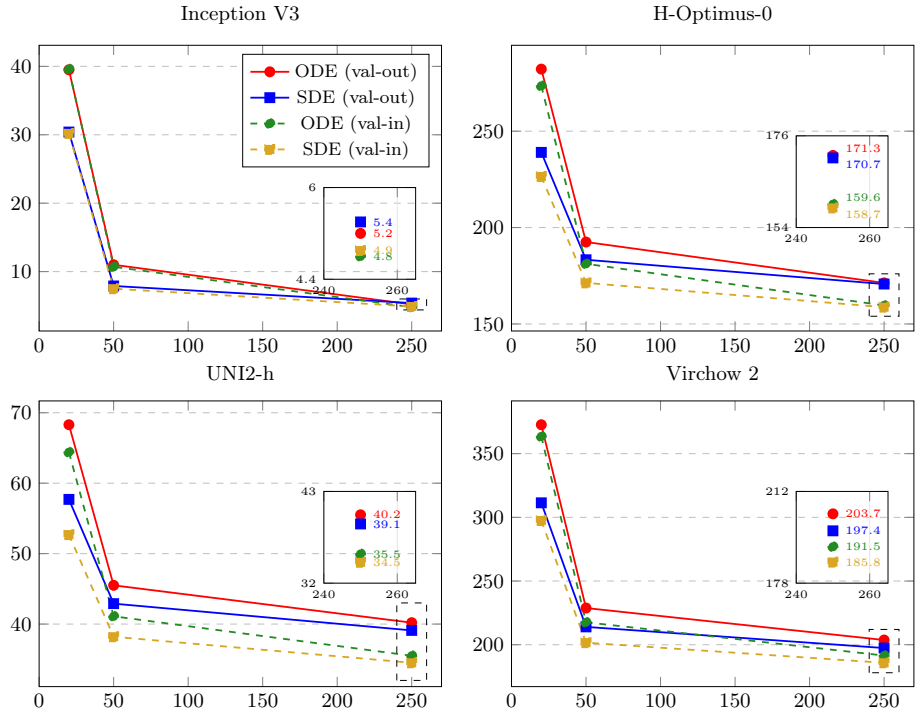

To obtain a more comprehensive evaluation of our models, we decided to compute several metrics in addition to the standard Fréchet Inception Distance ~\cite{heusel2017gans} (FID) and to use several state-of-the-art pathology-specific extractors (H-Optimus-0~\cite{hoptimus0}, UNI2-h~\cite{chen2024towards}, Virchow 2~\cite{zimmermann2024virchow2}, UNI~\cite{chen2024towards}, CONCH-v1~\cite{lu2024visual} and Phikon-v2~\cite{filiot2024phikon}) to compute them rather than relying solely on standard models like Inception-v3~\cite{szegedy2016rethinking} or DINOv2. Given the high fidelity of the generated images, we posit that pathology feature extractor will be able to uncover subtle differences in generated tiles that models trained on ImageNet-like datasets might miss. We use "FD" as the base metric name for the Fréchet distance computed with different extractors. Furthermore, we incorporated Feature Likelihood Divergence (FLD), recently introduced by Jiralerspong et al.~\cite{jiralerspong2023feature}, to account for novelty in addition to realism and diversity, and Precision and Recall~\cite{kynkaanniemi2019improved} to disentangle performance between coverage and sample realism. In addition, to precisely measure the quality of the learned conditioning and not only the overall realism, we used the H0-mini features of the validation sets as guidance to the diffusion model to create synthetic validation-like datasets. We then compared the cosine similarity of the embeddings between the original and synthetic sets sample-wise, with different extractors again. Finally, we investigated both ODE and SDE sampling with varying number of sampling steps in an unconditional sampling scenario for both validation sets (Figure \ref{fig:uncond_validation_grid}).\\

Based on our quantitative evaluation (Table \ref{tab:model_comparison}), we draw several conclusions:
\begin{itemize}
    \item \textbf{Data Scaling}: There is no gain in scaling the training data from 40M to 108M images, or in moving from a randomly sampled dataset to a thoroughly curated one, with the Fréchet distance increasing slightly across extractors. A similar observation has been made by Karasikov et al.~\cite{karasikov2025training} in the context of SSL models, where smaller training sets do not systematically translate to lower performance.
    \item \textbf{VAE EMA}: Computing an exponential moving average of the VAE weights to use for inference yielded observable quality improvements. While this improvement was not uniform across all metrics (e.g., the standard Inception-v3 FD slightly degraded in the EMA version), we observed consistent Fréchet Distance improvements across all histopathology-specific extractors. Consequently, we selected the EMA model for our subsequent experiments.
    \item \textbf{Metric Concordance}: We found an overall high model ranking agreement among the different metrics and across extractors (with FD computed with pathology extractors being the most sensitive). Precision and Recall metrics reached saturation, indicating good distribution coverage and realism but rendering them less discriminative for fine-grained model comparison. Conversely, the FLD score proved difficult to interpret, as model rankings fluctuated depending on the chosen feature extractor.
    \item \textbf{Overfitting Analysis}: Results obtained on the \textbf{val-in} dataset are consistently better than their \textbf{val-out} counterparts, suggesting a slide-level overfitting of our models. However, because a small but non-zero domain shift exists between the training subset used for guidance and the \textbf{val-out} dataset, the performance of our models must be weighted against this gap to assess overfitting. We computed the Fréchet Distance between the guidance subset and \textbf{val-in} and \textbf{val-out} as a baseline (row 4 of Table \ref{tab:model_comparison}). All the extractors detected a shift between the real conditioning subset and the \textbf{val-out} set, indicating that the performance degradation seen on \textbf{val-out} is partially attributable to this inherent distribution gap. Furthermore, because cosine similarity scores remained identical between \textbf{val-in} and \textbf{val-out} results, we conclude that our models do not significantly overfit to slide-level specificities.
    \item \textbf{Sampling Dynamics}: Our experiments indicate that ODE and SDE sampling schemes perform comparably after 250 sampling steps. However, SDE demonstrates a clear advantage at lower step counts. Additionally, while the most pronounced decrease in Fréchet Distance occurs between 20 and 50 steps, extending the process to 250 steps still yields measurable gains. From the Figure \ref{fig:uncond_validation_grid} and the Table \ref{tab:model_comparison}, we also note that conditional sampling achieves consistently better results than unconditional sampling, an observation aligned with prior research.
\end{itemize}

We release two models on HugginFace with the following naming convention: CytoSyn, corresponding to the model trained on the 40M dataset, and CytoSyn-v2 corresponding to the model trained on the 108M dataset with the EMA VAE. In addition, we release 100k synthetic tiles generated unconditionally with CytoSyn-v2.

\subsection{Out-of-distribution validation}

In addition to measuring our models' capacity to generate TCGA-like H\&E-stained tiles, we also investigated their ability to synthesize strongly out-of-distribution (OOD) samples. While unconditional sampling is inherently limited to the training distribution, conditional sampling can bypass this limitation by leveraging features from OOD tiles as conditioning (therefore relying on the robustness of the underlying feature extractor to guide the generation process). For this benchmark, we used data from the Study of a Prospective Adult Research Cohort with Inflammatory Bowel Disease~\cite{Sparcibd2021} (SPARC IBD) cohort, a multicentered longitudinal study of adult IBD patients. It provides both a non-oncology scenario, shifting the focus from the tumor microenvironments of TCGA to the inflammatory infiltrates and mucosal distortions characteristic of IBD, and new staining/scanner scenario as its slides and TCGA slides were digitized in different centers with different scanner brands (Olympus versus mainly Leica). SPARC IBD histology data consists of 3322 H\&E slides obtained from intestinal mucosal (mostly colon and ileum) biopsies of patients diagnosed with Crohn's disease, ulcerative colitis and other forms of IBD. We sampled 50k tiles uniformly at $\simeq20 \times$ magnification to use as conditioning and a distinct set of the same size as reference to compute the Fréchet distance.

\begin{table}
\centering
\caption{Comparison of FD score and cosine similarity of CytoSyn-v2 on the SPARC IBD tiles and on the \textbf{val-out} tiles, computed with different extractors.}\label{table:ood}
\begin{tabular}{|r|c|c|c|c|}
\hline
Performance metric $\downarrow$ & H-optimus-0 & Virchow 2 & UNI2-h & Inception V3 \\
\hline\hline
FD (SPARC-IBD) & 196.5 & 245.0 & 83.8 & 8.6 \\
FD (val-out) & 62.5 & 63.5 & 15.1 & 3.9 \\
\hline\hline
Cosine Sim. (SPARC-IBD) & 0.73 & 0.84 & 0.71 & 0.86 \\
Cosine Sim. (val-out) & 0.80 & 0.91 & 0.80 & 0.88 \\
\hline
\end{tabular}
\end{table}

Our OOD results (Table \ref{table:ood}) show that our model is sensitive to the distribution shift: we observed a noticeable FD increase consistent across extractors between the results on SPARC IBD and \textbf{val-out}. Cosine similarity followed a similar degradation trend. In contrast to our results, PixCell's experiments on their OOD dataset SPIDER~\cite{nechaev2025spider} yielded a near invariant Inception FD and a moderate increase for the other extractors, likely highlighting the benefit of having several sources in the training set. Nevertheless, in terms of absolute FD values, the out-of-distribution performance of our model reaches PixCell's in-distribution performance (Inception FD of around 8, Table \ref{tab:cytosyn_pixcell_comparison}). Further investigation is required to isolate the origin of our observed performance drop: whether it is driven by biologically relevant differences or a sensitivity to center-specific scanning and staining artifacts. Given that our training set already contains colonic histological patterns (via TCGA COAD tiles for instance), we posit that the latter is more likely.

\subsection{Comparison with PixCell}

To the best of our knowledge, this study represents the first instance where histopathology-specific diffusion models from different organizations are directly benchmarked together. Given the many differences between PixCell and our models, and to ensure the fairness of the comparison, we took into account some of the distinct design choices. First, we compared both models on the generation of TCGA tiles only (as TCGA is the intersection of their respective training distributions) rather than relying solely on published metrics derived from PixCell's own validation set (which is partly OOD for CytoSyn). Then, we focused our efforts on two impactful points in particular: 
\begin{itemize}
    \item \textbf{Image size}: PixCell generates $256 \times 256$ tiles conditioned on $256 \times 256$ tiles' features, whereas CytoSyn generates $224 \times 224$ tiles conditioned on $224 \times 224$ tiles' features. PixCell's guidance arm first resizes the images to $224 \times 224$ before inputting them to UNI2-h, while our guidance branch processes $224 \times 224$ images natively. This resizing operation slightly changes the resolution of the guidance tiles and introduces interpolation artifacts into the conditioning embeddings.
    \item \textbf{Image format}: PixCell's tiling pipeline is based on DS-MIL~\cite{li2021dual} which saves tiles in the JPEG format by default. An analysis of the PixCell repository confirms that the tiles were indeed likely saved as JPEG, while our pipeline extracts and saves tiles in the lossless PNG format. While extracting tiles as PNG files does not guarantee the complete absence of upstream compression artifacts, as JPEG compression can also be applied during the digitization of slides, it prevents additional compression loss. These artifacts will distort both the conditioning and the validation features used in the final metrics computation.
\end{itemize}

\begin{figure}[htbp]
    \centering
    
    \resizebox{\columnwidth}{!}{%
    \begin{tikzpicture}[
        node distance=0.6cm and 0.5cm,
        font=\sffamily\small\color{sciText},
        >={Stealth[length=2.5mm]},
        line/.style={draw=sciDraw, thick, ->},
        base/.style={
            rectangle, draw=sciDraw, align=center, rounded corners=2pt, 
            font=\sffamily\small,
            drop shadow={opacity=0.15, shadow xshift=1pt, shadow yshift=-1pt}
        },
        data/.style={
            base, fill=sciBlue, minimum width=2.4cm, minimum height=1.1cm
        },
        model/.style={
            base, fill=sciRed, minimum width=2.4cm, minimum height=1.1cm
        },
        op/.style={
            ellipse, draw=sciDraw, fill=sciGreen, align=center, 
            font=\sffamily\footnotesize, minimum width=1.8cm, minimum height=0.9cm,
            drop shadow={opacity=0.15, shadow xshift=1pt, shadow yshift=-1pt}
        },
        yellowop/.style={
            rectangle, draw=sciDraw, fill=sciYellow, align=center, 
            rounded corners=8pt, font=\sffamily\footnotesize,
            minimum width=1.8cm, minimum height=0.9cm,
            drop shadow={opacity=0.15, shadow xshift=1pt, shadow yshift=-1pt}
        },
        purpleop/.style={
            rectangle, draw=sciDraw, fill=sciPurple, align=center, 
            rounded corners=12pt, font=\sffamily\footnotesize,
            minimum width=1.8cm, minimum height=0.9cm,
            drop shadow={opacity=0.15, shadow xshift=1pt, shadow yshift=-1pt}
        },
        metric/.style={
            rectangle, draw=sciDraw, fill=white, rounded corners=10pt, 
            align=center, font=\sffamily\footnotesize\bfseries,
            minimum width=1.4cm, minimum height=0.7cm,
            drop shadow={opacity=0.15}
        },
        stack/.style={
            rectangle, draw=sciDraw!50, fill=white, 
            minimum width=2.4cm, minimum height=1.1cm, rounded corners=2pt
        }
    ]


    \node[model] (uni2h) at (0, 0) {UNI-2h / H0-mini};
    \node[model, below=0.8cm of uni2h] (pixcell) {PixCell / CytoSyn};
    \draw[line] (uni2h) -- (pixcell);


    \node[op, left=of uni2h] (rs1) {Resize\\$224 \times 224$};
    \node[purpleop, left=of rs1] (jpg1) {JPEG compression \\ (PixCell only)};
    \node[yellowop, left=of jpg1] (cc1) {Center-Crop\\$224 \times 224$\\(Cytosyn only)};
    
    \coordinate (cond_pos) at ($(cc1)+(-2.8, 0)$);
    \foreach \x in {3, 2, 1} { \node[stack, shift={(-\x*1.5pt, \x*1.5pt)}] at (cond_pos) {}; }
    \node[data] (cond) at (cond_pos) {Cond. Tiles\\$256 \times 256$};

    \draw[line] (cond) -- (cc1);
    \draw[line] (cc1) -- (jpg1);
    \draw[line] (jpg1) -- (rs1);
    \draw[line] (rs1) -- (uni2h);


    \coordinate (synth_pos) at ($(cond_pos)+(0, -3.4)$);
    \foreach \x in {3, 2, 1} { \node[stack, shift={(-\x*1.5pt, \x*1.5pt)}] at (synth_pos) {}; }
    \node[data] (synth) at (synth_pos) {Synth. Tiles};

    \draw[line] (pixcell.west) -| (synth.north);

    \node[op] (rs_synth) at (jpg1 |- synth) {Resize\\$224 \times 224$};
    \draw[line] (synth) -- (rs_synth);
    
    \node[model] (dino_top) at (uni2h |- synth) {Feature Extractor};

    \draw[line] (rs_synth) -- (dino_top);


    \node[model, below=1.2cm of dino_top] (dino_bot) {Feature Extractor};

    \node[op] (rs3) at (rs1 |- dino_bot) {Resize\\$224 \times 224$};
    \node[purpleop] (jpg2) at (jpg1 |- dino_bot) {JPEG compression \\ (PixCell only)};
    \node[yellowop] (cc2) at (cc1 |- dino_bot) {Center-Crop\\$224 \times 224$ \\(CytoSyn only)};

    \coordinate (test_pos) at (cond |- dino_bot);
    \foreach \x in {3, 2, 1} { \node[stack, shift={(-\x*1.5pt, \x*1.5pt)}] at (test_pos) {}; }
    \node[data] (test) at (test_pos) {Val. Tiles\\$256 \times 256$};

    \draw[line] (test) -- (cc2);
    \draw[line] (cc2) -- (jpg2);
    \draw[line] (jpg2) -- (rs3);
    \draw[line] (rs3) -- (dino_bot);

    
    \path (dino_top) -- node[metric] (fid) {FID \& other metrics} (dino_bot);
    \draw[line] (dino_top) -- (fid);
    \draw[line] (dino_bot) -- (fid);

    \end{tikzpicture}%
    }
    \caption{Overview of our all-in-one validation pipeline.}
    \label{fig:validation_pipeline}
\end{figure}
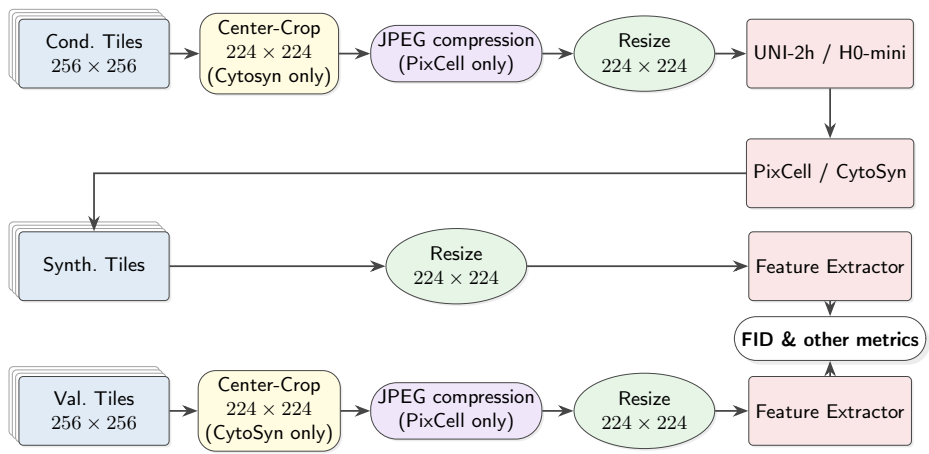

We first applied CytoSyn's original validation pipeline (equivalent to the pipeline in Figure \ref{fig:validation_pipeline}, with a center-crop operation for both models and no JPEG compression) on images generated with PixCell. We did not compute all metrics for this scenario. Then, to account for the differences between the models, we performed a step-wise ablation: 

\begin{itemize}
\item \textbf{Image Size Adjustment}: We modified our \textbf{val-out} dataset by expanding the original tile coordinates by $\pm 16$ pixels, enlarging the tiles to $256 \times 256$. Consequently, the original validation dataset becomes a center-cropped version of this new $256 \times 256$ dataset. We performed the same transformation for the conditioning subset. The UNI2-h features computed on this new dataset were then used as inputs for the conditional sampling.
\item \textbf{Validation JPEG Compression }: To mimic the validation data used for PixCell, which likely contained JPEG artifacts, we created a JPEG version of the $256 \times 256 $ \textbf{val-out} dataset, with a JPEG quality of 70 (the DS-MIL default), and keep the $256 \times 256$ guidance subset in its previous PNG version. 
\item \textbf{Conditioning JPEG Compression}: To further understand the effect of compression, we created a JPEG version of the $256 \times 256$ guidance subset, and recomputed the conditioning UNI2-h features again.
\end{itemize}

Splitting the JPEG experiments into two steps allowed us to disentangle the origin of the remaining performance gap after taking into account image size: whether it arose from JPEG artifacts in the generated images or JPEG artifact effects in the conditioning features. A complete overview of the final validation pipeline is available in Figure \ref{fig:validation_pipeline}. As a negative control, we also computed a FD score using our model's images and a JPEG-compressed version of the $224 \times 224$ \textbf{val-out} dataset. This ensured that the FD decrease observed with PixCell was not a general effect of the JPEG-compression. Finally, we note that discrepancies beyond image size and compression remain, for instance, PixCell utilized the Clean-FID implementation, whereas we relied on the Jiralerspong et al. implementation. Differences in underlying resizing operations and interpolation kernels are known to affect FID scores, and we leave this additional investigation to future work.\\

To align with CytoSyn’s inference configuration, our preliminary experiments evaluated PixCell using 250 sampling steps rather than the 50 steps utilized in the original study. However, upon observing negligible differences in image quality between the 50-step and 250-step regimes, we reverted to 50 steps to accelerate the evaluation process. Furthermore, because PixCell was trained on a highly heterogeneous dataset encompassing multiple data sources, its unconditionally generated images naturally reflect this broader distribution. Because our validation set is strictly derived from the TCGA cohort, unconditional generation metrics would be artificially penalized by this domain mismatch. Consequently, only conditional sampling metrics provide a meaningful comparison and are reported here. Finally, we note that because TCGA data is a core component of PixCell's training data, this benchmark effectively serves as a rigorous in-distribution validation for their model. To complement this in-distribution evaluation, we also conducted an out-of-distribution benchmark using the previously described SPARC IBD dataset. For this OOD comparison, we applied directly the final validation pipeline (with both the $256 \times 256$ tiles and the JPEG compression for validation and guidance images).\\

\begin{table}[h!]
\centering
\fontsize{8}{\baselineskip}\selectfont 
\caption{FD and cosine similarity for CytoSyn-v2 and PixCell on the \textbf{val-out} and SPARC IBD datasets across different scenarios. Approximate PixCell results were read from the paper's figures, while precise cosine similarities were obtained from the arXiv v1 version. PixCell results were obtained with a classifier-free guidance scale of 2.0. All images generated with PixCell and CytoSyn-v2 were saved as PNGs.}
\label{tab:cytosyn_pixcell_comparison}
\begin{tabularx}{\textwidth}{|X|c|c|c|c|}
\hline
\multicolumn{5}{|c|}{\textbf{FD - \textbf{val-out}}} \\
\hline
Model \& validation details $\downarrow$ & H-Optimus-0 & Virchow 2 & UNI2-h & Inception V3 \\ \hline
PixCell & & & & \\
Original paper's in-domain results & -- & $\sim 140$ & -- & $\sim 8$ \\
224px + PNG images (all) - 250 steps & -- & -- & -- & 61.5 \\ 
256px + PNG images (all) - 250 steps & 355.9 & 368.0 & 95.6 & 28.5 \\
256px + PNG images (all) - 50 steps & 346.2 & 368.4 & 94.4 & 29.0 \\
256px + JPEG \textbf{val-out} only - 250 steps & 210.5 & 257.3 & 58.3 & 10.1 \\
256px + JPEG \textbf{val-out} only - 50 steps & 207.8 & 266.6 & 58.4 & 10.4 \\
256px + JPEG images (all) - 50 steps & 194.3 & 206.1 & 48.0 & 5.5 \\\hline
CytoSyn-v2 (JPEG \textbf{val-out}) & 212.4 & 168.2 & 76.3 & 40.4 \\
CytoSyn-v2 & \textbf{62.5} & \textbf{63.5} & \textbf{15.1} & \textbf{3.9 }\\ \hline\hline
\multicolumn{5}{|c|}{\textbf{FD - SPARC IBD}} \\ \hline
PixCell - 256px, JPEG images, 50 steps  & 550.7 & 668.1 & 340.5 & 26.7 \\
CytoSyn-v2 & 196.5 & 245.0 & 83.8 & 8.6 \\\hline\hline
\multicolumn{5}{|c|}{\textbf{Cosine Similarity - \textbf{val-out}}} \\ \hline
Model \& validation details $\downarrow$ & UNI & CONCH-v1 & Phikon-v2 & Virchow 2 \\ \hline
PixCell & & & & \\
Original paper's in-domain results & 0.70 & 0.89 & \textbf{0.83} & $\sim 0.8$\\
256px + PNG images (all) - 250 steps & 0.54 & 0.75 & 0.45 & 0.72 \\
256px + JPEG \textbf{val-out} only - 250 steps& 0.64 & 0.81 & 0.75 & 0.75 \\
256px + JPEG images (all) - 50 steps & 0.70 & 0.84 & 0.81 & 0.79 \\\hline
CytoSyn-v2 & \textbf{0.80} &\textbf{ 0.91} & 0.81 & \textbf{0.91} \\ \hline\hline
\multicolumn{5}{|c|}{\textbf{Cosine Similarity - SPARC IBD}} \\ \hline
Pixcell - 256px, JPEG images, 50 steps & 0.49 & 0.72 & 0.71 & 0.63 \\
CytoSyn-v2 & \textbf{0.76} & \textbf{0.86} & \textbf{0.70} & \textbf{0.84}\\\hline
\end{tabularx}
\end{table}

Our results first highlight the extreme sensitivity of both diffusion models and performance metrics to mundane preprocessing pipeline details. Indeed, accounting for image size and format, we were able to reduce PixCell's Inception FD score by an order of magnitude (from 61.5 to 5.5). While the exact magnitude varied, this behavior was consistent across different extractors for both FD and embeddings similarity. We found that PixCell learned JPEG artifacts in two distinct areas during training: in generated images, and in the conditioning pathway. Indeed, utilizing guidance features computed from JPEG tiles in addition to a JPEG validation set consistently improved PixCell's results across all metrics. Our negative control (CytoSyn-v2 + JPEG \textbf{val-out}) confirmed that this performance boost seems specifically tied to PixCell's training pipeline and is not a universal feature-level effect of JPEG compression. In our experiments, we managed to reproduce results close to the original PixCell paper scores, particularly for the cosine similarity. We therefore posit that the initial reproducibility gap was primarily driven by discrepancies in image size and file format. The remaining inconsistencies (e.g., our reproduced Inception FD being lower than PixCell's reported results, while the Virchow 2 FD was higher) may stem from differences of validation sets (PixCell in-domain validation set incorporates data from multiple sources beyond TCGA) or finer pipeline differences (such as the aforementioned metric and resize implementation, the use of mixed-precision, etc).\\

After accounting for differences in data preparation, CytoSyn-v2 consistently outperforms PixCell in generation quality, whether evaluated on the TCGA validation set with reproduced results or compared directly against PixCell's originally published metrics. This advantage is confirmed on the SPARC IBD cohort. While both models exhibit a noticeable performance drop in this OOD scenario, our results demonstrate the superior robustness of our model, seen across metrics and extractors (e.g., an Inception FD of 8.6 compared to PixCell's 26.7). These findings do not align with the good generalization capabilities observed on PixCell's own OOD benchmark on the SPIDER~\cite{nechaev2025spider} dataset, and further investigation is required to understand this apparently contradictory behavior (possible reasons include different scanner brands, staining protocols, slightly different MPP, etc.). Given that our model was trained exclusively on TCGA diagnostic slides, we attribute its robustness to H0-mini. Indeed, this model stands among the most robust histology feature extractors currently available~\cite{filiot2025distilling,komen2025towards,thiringer2026scanner}, and constraining the diffusion model's latent space to align with H0-mini's embeddings likely transferred the extractor's broad generalization capabilities directly to the generative model.

\section{A note on variability}

The implementation by Jiralerspong et al.~\cite{jiralerspong2023feature} of the FID score restricts the number of synthetic samples to 50k while keeping the entire real set. Other implementations~\cite{parmar2022aliased} follow this strategy as well. To account for stochasticity in CytoSyn's inference and tile selection and obtain a standard deviation for our results, we used a bootstrapping procedure (sampling of 50k synthetic tiles from the 100k pool 50 times). We initially performed this analysis for the Fréchet distance across all extractors using CytoSyn. Upon observation that results did not fluctuate significantly (Table \ref{tab:variance_ci}), and that a similar conclusion was reached independently by PixCell, we did not conduct this analysis for subsequent experiments. All results in Table \ref{tab:model_comparison} and Table \ref{tab:cytosyn_pixcell_comparison} were obtained with the same seed for the synthetic samples selection.

\begin{table}[h!]
\centering
\caption{CytoSyn's Mean $\pm$ standard deviation obtained with bootstrapping for the Fréchet distance metric, across validation sets and extractors.}
\label{tab:variance_ci}
\begin{tabular}{|c|c|c|c|c|}
\cline{2-5}
\multicolumn{1}{c|}{} & H-Optimus-0 & Virchow-2 & UNI2-h & Inception-v3 \\ \hline
val-out & $72.17 \pm 0.18$ & $70.23 \pm 0.42$ & $16.67 \pm 0.05$ & $3.40 \pm 0.03$ \\
val-in & $58.27 \pm 0.18$ & $55.41 \pm 0.45$ & $10.94 \pm 0.04$ & $2.90 \pm 0.02$ \\
\hline
\end{tabular}
\end{table}

\section{Conclusion}

In this work, we introduced CytoSyn, a novel family of foundation diffusion models tailored specifically to histopathology. Outperforming current baselines, our models achieve state-of-the-art results in generating H\&E-stained tiles and demonstrate strong out-of-distribution generalization on an unseen clinical indication. Beyond confirming high synthesis quality, we conducted an exploration of different methodological choices regarding both the diffusion models and the benchmarking process and investigated several properties of pathology diffusion models, such as the slide-level overfitting tendency and the out-of-distribution behavior. Through a rigorous comparison with PixCell, our study sheds light on the important sensitivity of generative models and evaluation metrics to seemingly trivial technical choices, such as image resizing and compression. Whole-slide image processing pipelines are complex and the downstream impact of their many details is rarely quantified. This work underscores their importance and highlights ongoing reproducibility challenges in the field. We publicly released our models and additional data to encourage the pathology research community to further investigate the potential of domain-specific generative foundation models.

\section*{Acknowledgment}

This work was granted access to the High Performance Computing (HPC) resources of Meluxina, from LuxProvide, as part of a Euro-HPC grant under the allocation EHPC-AI-2024A04-020, and to the HPC resources of IDRIS under the allocations 2025-A0181012519 made by GENCI. The results published here are in part based on data and biosamples obtained from the IBD Plexus program of the Crohn’s \& Colitis Foundation and in part based upon data generated by
the TCGA Research Network: \href{https://www.cancer.gov/tcga}{https://www.cancer.gov/tcga}.

%
%
%
\bibliographystyle{splncs04}
\bibliography{biblio}

@inproceedings{sohl2015deep,
  title={Deep unsupervised learning using nonequilibrium thermodynamics},
  author={Sohl-Dickstein, Jascha and Weiss, Eric and Maheswaranathan, Niru and Ganguli, Surya},
  booktitle={International conference on machine learning},
  year={2015},
}

@article{song2019generative,
  title={Generative modeling by estimating gradients of the data distribution},
  author={Song, Yang and Ermon, Stefano},
  journal={Advances in neural information processing systems},
  year={2019}
}

@article{ho2020denoising,
  title={Denoising diffusion probabilistic models},
  author={Ho, Jonathan and Jain, Ajay and Abbeel, Pieter},
  journal={Advances in neural information processing systems},
  year={2020}
}

@article{song2020score,
  title={Score-based generative modeling through stochastic differential equations},
  author={Song, Yang and Sohl-Dickstein, Jascha and Kingma, Diederik P and Kumar, Abhishek and Ermon, Stefano and Poole, Ben},
  journal={arXiv preprint arXiv:2011.13456},
  year={2020}
}

@inproceedings{ldm,
  title={High-resolution image synthesis with latent diffusion models},
  author={Rombach, Robin and Blattmann, Andreas and Lorenz, Dominik and Esser, Patrick and Ommer, Bj{\"o}rn},
  booktitle={Proceedings of the IEEE/CVF conference on computer vision and pattern recognition},
  year={2022}
}

@article{dhariwal2021diffusion,
  title={{Diffusion models beat GANs on image synthesis}},
  author={Dhariwal, Prafulla and Nichol, Alexander},
  journal={Advances in neural information processing systems},
  year={2021}
}

@inproceedings{repa,
  title={Representation alignment for generation: Training diffusion transformers is easier than you think},
  author={Yu, Sihyun and Kwak, Sangkyung and Jang, Huiwon and Jeong, Jongheon and Huang, Jonathan and Shin, Jinwoo and Xie, Saining},
  booktitle={13th International Conference on Learning Representations, ICLR 2025},
  year={2025},
}

@inproceedings{repae,
  title={Repa-e: Unlocking vae for end-to-end tuning of latent diffusion transformers},
  author={Leng, Xingjian and Singh, Jaskirat and Hou, Yunzhong and Xing, Zhenchang and Xie, Saining and Zheng, Liang},
  booktitle={Proceedings of the IEEE/CVF International Conference on Computer Vision},
  year={2025}
}

@inproceedings{sit,
  title={Sit: Exploring flow and diffusion-based generative models with scalable interpolant transformers},
  author={Ma, Nanye and Goldstein, Mark and Albergo, Michael S and Boffi, Nicholas M and Vanden-Eijnden, Eric and Xie, Saining},
  booktitle={European Conference on Computer Vision},
  year={2024},
}

@inproceedings{dit,
  title={Scalable diffusion models with transformers},
  author={Peebles, William and Xie, Saining},
  booktitle={Proceedings of the IEEE/CVF international conference on computer vision},
  year={2023}
}

@article{lipman2022flow,
  title={Flow matching for generative modeling},
  author={Lipman, Yaron and Chen, Ricky TQ and Ben-Hamu, Heli and Nickel, Maximilian and Le, Matt},
  journal={arXiv preprint arXiv:2210.02747},
  year={2022}
}

@article{liu2022flow,
  title={Flow straight and fast: Learning to generate and transfer data with rectified flow},
  author={Liu, Xingchao and Gong, Chengyue and Liu, Qiang},
  journal={arXiv preprint arXiv:2209.03003},
  year={2022}
}

@inproceedings{
    gao2025diffusionmeetsflow,
    title={Diffusion Models and Gaussian Flow Matching: Two Sides of the Same Coin},
    author={Ruiqi Gao and Emiel Hoogeboom and Jonathan Heek and Valentin De Bortoli and Kevin Patrick Murphy and Tim Salimans},
    booktitle={The Fourth Blogpost Track at ICLR 2025},
    year={2025},
    url={https://openreview.net/forum?id=C8Yyg9wy0s}
}

@article{albergo2023stochastic,
  title={Stochastic interpolants: A unifying framework for flows and diffusions},
  author={Albergo, Michael S and Boffi, Nicholas M and Vanden-Eijnden, Eric},
  journal={arXiv preprint arXiv:2303.08797},
  year={2023}
}

@article{yellapragada2025pixcell,
  title={PixCell: A generative foundation model for digital histopathology images},
  author={Yellapragada, Srikar and Graikos, Alexandros and Li, Zilinghan and Triaridis, Kostas and Belagali, Varun and Kapse, Saarthak and Nandi, Tarak Nath and Madduri, Ravi K and Prasanna, Prateek and others},
  journal={arXiv preprint arXiv:2506.05127},
  year={2025}
}

@misc{histoplus2025,
  title        = {Towards Comprehensive Cellular Characterisation of H\&E Slides},
  author       = {B. Adjadj and P.-A. Bannier and G. Horent and S. Mandela and A. Lyon and K. Schutte and U. Marteau and V. Gaury and L. Dumont and T. Mathieu and R. Belbahri and B. Schmauch and E. Durand and K. Von Loga and L. Gillet},
  year         = {2025},
  journal          = {arXiv preprint arXiv:2508.09926}
}

@inproceedings{filiot2025distilling,
  title={Distilling foundation models for robust and efficient models in digital pathology},
  author={Filiot, Alexandre and Dop, Nicolas and Tchita, Oussama and Riou, Auriane and Dubois, R{\'e}my and Peeters, Thomas and Valter, Daria and Scalbert, Marin and Saillard, Charlie and Robin, Genevi{\`e}ve and others},
  booktitle={International Conference on Medical Image Computing and Computer-Assisted Intervention},
  year={2025},
}

@article{Sparcibd2021,
    author = {Raffals, Laura E and Saha, Sumona and Bewtra, Meenakshi and Norris, Cecile and Dobes, Angela and Heller, Caren and O’Charoen, Sirimon and Fehlmann, Tara and others},
    title = {The Development and Initial Findings of A Study of a Prospective Adult Research Cohort with Inflammatory Bowel Disease (SPARC IBD)},
    journal = {Inflammatory Bowel Diseases},
    year = {2021},
}

@article{weinstein2013cancer,
  title={The cancer genome atlas pan-cancer analysis project},
  author={Weinstein, John N and Collisson, Eric A and Mills, Gordon B and Shaw, Kenna R and Ozenberger, Brad A and Ellrott, Kyle and Shmulevich, Ilya and Sander, Chris and Stuart, Joshua M},
  journal={Nature genetics},
  year={2013},
}

@software{hoptimus0,
  author = {Saillard, Charlie and Jenatton, Rodolphe and Llinares-López, Felipe and Mariet, Zelda and Cahané, David and Durand, Eric and Vert, Jean-Philippe},
  title = {H-optimus-0},
  url = {https://github.com/bioptimus/releases/tree/main/models/h-optimus/v0},
  year = {2024},
}

@article{oquab2024dinov2,
  title={DINOv2: Learning Robust Visual Features without Supervision},
  author={Oquab, Maxime and Darcet, Timoth{\'e}e and Moutakanni, Th{\'e}o and Vo, Huy and Szafraniec, Marc and Khalidov, Vasil and Fernandez, Pierre and Haziza, Daniel and Massa, Francisco and El-Nouby, Alaaeldin and others},
  journal={Transactions on Machine Learning Research Journal},
  year={2024}
}

@inproceedings{szegedy2016rethinking,
  title={Rethinking the inception architecture for computer vision},
  author={Szegedy, Christian and Vanhoucke, Vincent and Ioffe, Sergey and Shlens, Jon and Wojna, Zbigniew},
  booktitle={Proceedings of the IEEE conference on computer vision and pattern recognition},
  year={2016}
}

@inproceedings{parmar2022aliased,
  title={On aliased resizing and surprising subtleties in gan evaluation},
  author={Parmar, Gaurav and Zhang, Richard and Zhu, Jun-Yan},
  booktitle={Proceedings of the IEEE/CVF conference on computer vision and pattern recognition},
  year={2022}
}

@article{edwards2015cptac,
  title={The CPTAC data portal: a resource for cancer proteomics research},
  author={Edwards, Nathan J and Oberti, Mauricio and Thangudu, Ratna R and Cai, Shuang and McGarvey, Peter B and Jacob, Shine and Madhavan, Subha and Ketchum, Karen A},
  journal={Journal of proteome research},
  year={2015},
}

@article{lonsdale2013genotype,
  title={The genotype-tissue expression (GTEx) project},
  author={Lonsdale, John and Thomas, Jeffrey and Salvatore, Mike and Phillips, Rebecca and Lo, Edmund and Shad, Saboor and Hasz, Richard and Walters, Gary and Garcia, Fernando and Young, Nancy and others},
  journal={Nature genetics},
  year={2013},
}

@article{jiralerspong2023feature,
  title={Feature likelihood divergence: evaluating the generalization of generative models using samples},
  author={Jiralerspong, Marco and Bose, Joey and Gemp, Ian and Qin, Chongli and Bachrach, Yoram and Gidel, Gauthier},
  journal={Advances in Neural Information Processing Systems},
  year={2023}
}

@article{chen2024towards,
  title={Towards a general-purpose foundation model for computational pathology},
  author={Chen, Richard J and Ding, Tong and Lu, Ming Y and Williamson, Drew FK and Jaume, Guillaume and Song, Andrew H and Chen, Bowen and Zhang, Andrew and Shao, Daniel and Shaban, Muhammad and others},
  journal={Nature medicine},
  year={2024},
}

@article{zimmermann2024virchow2,
  title={Virchow2: Scaling self-supervised mixed magnification models in pathology},
  author={Zimmermann, Eric and Vorontsov, Eugene and Viret, Julian and Casson, Adam and Zelechowski, Michal and Shaikovski, George and Tenenholtz, Neil and Hall, James and Klimstra, David and Yousfi, Razik and others},
  journal={arXiv preprint arXiv:2408.00738},
  year={2024}
}

@article{filiot2024phikon,
  title={Phikon-v2, a large and public feature extractor for biomarker prediction},
  author={Filiot, Alexandre and Jacob, Paul and Mac Kain, Alice and Saillard, Charlie},
  journal={arXiv preprint arXiv:2409.09173},
  year={2024}
}

@article{lu2024visual,
  title={A visual-language foundation model for computational pathology},
  author={Lu, Ming Y and Chen, Bowen and Williamson, Drew FK and Chen, Richard J and Liang, Ivy and Ding, Tong and Jaume, Guillaume and Odintsov, Igor and Le, Long Phi and Gerber, Georg and others},
  journal={Nature medicine},
  year={2024},
}

@inproceedings{ho2021classifier,
  title={Classifier-Free Diffusion Guidance},
  author={Ho, Jonathan and Salimans, Tim},
  booktitle={NeurIPS 2021 Workshop on Deep Generative Models and Downstream Applications},
  year={2021}
}

@inproceedings{esser2024scaling,
  title={Scaling Rectified Flow Transformers for High-Resolution Image Synthesis},
  author={Esser, Patrick and Kulal, Sumith and Blattmann, Andreas and Entezari, Rahim and M{\"u}ller, Jonas and Saini, Harry and Levi, Yam and Lorenz, Dominik and Sauer, Axel and others},
  booktitle={International Conference on Machine Learning},
  year={2024},
}

@article{heusel2017gans,
  title={Gans trained by a two time-scale update rule converge to a local nash equilibrium},
  author={Heusel, Martin and Ramsauer, Hubert and Unterthiner, Thomas and Nessler, Bernhard and Hochreiter, Sepp},
  journal={Advances in neural information processing systems},
  year={2017}
}

@article{kynkaanniemi2019improved,
  title={Improved precision and recall metric for assessing generative models},
  author={Kynk{\"a}{\"a}nniemi, Tuomas and Karras, Tero and Laine, Samuli and Lehtinen, Jaakko and Aila, Timo},
  journal={Advances in neural information processing systems},
  year={2019}
}

@article{komen2025towards,
  title={Towards robust foundation models for digital pathology},
  author={K{\"o}men, Jonah and de Jong, Edwin D and Hense, Julius and Marienwald, Hannah and Dippel, Jonas and Naumann, Philip and Marcus, Eric and Ruff, Lukas and Alber, Maximilian and Teuwen, Jonas and others},
  journal={arXiv preprint arXiv:2507.17845},
  year={2025}
}

@article{thiringer2026scanner,
  title={Scanner-Induced Domain Shifts Undermine the Robustness of Pathology Foundation Models},
  author={Thiringer, Erik and Gustafsson, Fredrik K and Eriksson, Kajsa Ledesma and Rantalainen, Mattias},
  journal={arXiv preprint arXiv:2601.04163},
  year={2026}
}

@article{jaume2024hest,
  title={Hest-1k: A dataset for spatial transcriptomics and histology image analysis},
  author={Jaume, Guillaume and Doucet, Paul and Song, Andrew and Lu, Ming Yang and Almagro P{\'e}rez, Cristina and Wagner, Sophia and Vaidya, Anurag and Chen, Richard and Williamson, Drew and Kim, Ahrong and others},
  journal={Advances in Neural Information Processing Systems},
  year={2024}
}

@article{neidlinger2025benchmarking,
  title={Benchmarking foundation models as feature extractors for weakly supervised computational pathology},
  author={Neidlinger, Peter and El Nahhas, Omar SM and Muti, Hannah Sophie and Lenz, Tim and Hoffmeister, Michael and Brenner, Hermann and van Treeck, Marko and Langer, Rupert and Dislich, Bastian and Behrens, Hans Michael and others},
  journal={Nature biomedical engineering},
  year={2025},
}

@inproceedings{gatopoulos2024eva,
  title={eva: Evaluation framework for pathology foundation models},
  author={Gatopoulos, Ioannis and K{\"a}nzig, Nicolas and Moser, Roman and Ot{\'a}lora, Sebastian and others},
  booktitle={Medical Imaging with Deep Learning},
  year={2024}
}

@article{campanella2025clinical,
  title={A clinical benchmark of public self-supervised pathology foundation models},
  author={Campanella, Gabriele and Chen, Shengjia and Singh, Manbir and Verma, Ruchika and Muehlstedt, Silke and Zeng, Jennifer and Stock, Aryeh and Croken, Matt and Veremis, Brandon and Elmas, Abdulkadir and others},
  journal={Nature Communications},
  year={2025},
}

@inproceedings{tsai2024test,
  title={Test-time stain adaptation with diffusion models for histopathology image classification},
  author={Tsai, Cheng-Chang and Chen, Yuan-Chih and Lu, Chun-Shien},
  booktitle={European Conference on Computer Vision},
  year={2024},
}

@inproceedings{graikos2024learned,
  title={Learned representation-guided diffusion models for large-image generation},
  author={Graikos, Alexandros and Yellapragada, Srikar and Le, Minh-Quan and Kapse, Saarthak and Prasanna, Prateek and Saltz, Joel and Samaras, Dimitris},
  booktitle={Proceedings of the IEEE/CVF Conference on Computer Vision and Pattern Recognition},
  year={2024}
}

@article{belagali2024gen,
  title={Gen-sis: Generative self-augmentation improves self-supervised learning},
  author={Belagali, Varun and Yellapragada, Srikar and Graikos, Alexandros and Kapse, Saarthak and Li, Zilinghan and Nandi, Tarak Nath and Madduri, Ravi K and Prasanna, Prateek and Saltz, Joel and Samaras, Dimitris},
  journal={arXiv preprint arXiv:2412.01672},
  year={2024}
}

@InProceedings{Yellapragada_2025_CVPR,
  author = {Yellapragada, Srikar and Graikos, Alexandros and Triaridis, Kostas and Prasanna, Prateek and Gupta, Rajarsi and Saltz, Joel and Samaras, Dimitris},
  title = {ZoomLDM: Latent Diffusion Model for Multi-scale Image Generation},
  booktitle = {Proceedings of the Computer Vision and Pattern Recognition Conference (CVPR)},
  year = {2025},
}

@article{vzigutyte2025counterfactual,
  title={Counterfactual Diffusion Models for Interpretable Morphology-based Explanations of Artificial Intelligence Models in Pathology},
  author={{\v{Z}}igutyt{\.e}, Laura and Lenz, Tim and Han, Tianyu and Hewitt, Katherine J and Reitsam, Nic G and Foersch, Sebastian and Carrero, Zunamys I and Unger, Michaela and Pearson, Alexander T and Truhn, Daniel and others},
  journal={bioRxiv},
  year={2025}
}

@article{carrillo2025generation,
  title={Generation of synthetic whole-slide image tiles of tumours from RNA-sequencing data via cascaded diffusion models},
  author={Carrillo-Perez, Francisco and Pizurica, Marija and Zheng, Yuanning and Nandi, Tarak Nath and Madduri, Ravi and Shen, Jeanne and Gevaert, Olivier},
  journal={Nature Biomedical Engineering},
  volume={9},
  number={3},
  pages={320--332},
  year={2025},
  publisher={Nature Publishing Group UK London}
}

@inproceedings{zhou2022image,
    title={Image {BERT} Pre-training with Online Tokenizer},
    author={Jinghao Zhou and Chen Wei and Huiyu Wang and Wei Shen and Cihang Xie and Alan Yuille and Tao Kong},
    booktitle={International Conference on Learning Representations},
    year={2022},
}

@inproceedings{shen2022randstainna,
  title={Randstainna: Learning stain-agnostic features from histology slides by bridging stain augmentation and normalization},
  author={Shen, Yiqing and Luo, Yulin and Shen, Dinggang and Ke, Jing},
  booktitle={International Conference on Medical Image Computing and Computer-Assisted Intervention},
  pages={212--221},
  year={2022},
  organization={Springer}
}

@article{nechaev2025spider,
  title={SPIDER: a comprehensive multi-organ supervised pathology dataset and baseline models},
  author={Nechaev, Dmitry and Pchelnikov, Alexey and Ivanova, Ekaterina},
  journal={arXiv preprint arXiv:2503.02876},
  year={2025}
}

@inproceedings{li2021dual,
  title={Dual-stream multiple instance learning network for whole slide image classification with self-supervised contrastive learning},
  author={Li, Bin and Li, Yin and Eliceiri, Kevin W},
  booktitle={Proceedings of the IEEE/CVF conference on computer vision and pattern recognition},
  year={2021}
}

@inproceedings{karasikov2025training,
  title={Training state-of-the-art pathology foundation models with orders of magnitude less data},
  author={Karasikov, Mikhail and van Doorn, Joost and K{\"a}nzig, Nicolas and Erdal Cesur, Melis and Horlings, Hugo Mark and Berke, Robert and Tang, Fei and Ot{\'a}lora, Sebastian},
  booktitle={International Conference on Medical Image Computing and Computer-Assisted Intervention},
  pages={573--583},
  year={2025},
  organization={Springer}
}

\end{document}